\documentclass[conference]{IEEEtran}
\IEEEoverridecommandlockouts

\usepackage[numbers]{natbib}
\usepackage{multicol}
\usepackage[bookmarks=true]{hyperref}

\usepackage{times}
\usepackage{epsfig}
\usepackage{graphicx}
\usepackage{amsmath}
\usepackage{amssymb}

\usepackage{mathtools}

\usepackage{tikz}

\newcommand{\beq}{\begin{equation}}
\newcommand{\eeq}{\end{equation}}
\newcommand{\bdm}{\begin{displaymath}}
\newcommand{\edm}{\end{displaymath}}
\newcommand{\bea}{\begin{eqnarray}}
\newcommand{\eea}{\end{eqnarray}}
\newcommand{\beas}{\begin{eqnarray*}}
\newcommand{\eeas}{\end{eqnarray*}}
\newcommand{\ba}{\begin{array}}
\newcommand{\ea}{\end{array}}
\newcommand{\bit}{\begin{itemize}}
\newcommand{\eit}{\end{itemize}}
\newcommand{\ben}{\begin{enumerate}}
\newcommand{\een}{\end{enumerate}}

\usepackage{subfigure}

\newcommand\copyrighttext{%
  \footnotesize \textcopyright 2016 IEEE. Personal use of this material is permitted.
  Permission from IEEE must be obtained for all other uses, in any current or future
  media, including reprinting/republishing this material for advertising or promotional
  purposes, creating new collective works, for resale or redistribution to servers or
  lists, or reuse of any copyrighted component of this work in other works.
}
\newcommand\copyrightnotice{%
\begin{tikzpicture}[remember picture,overlay]
\node[anchor=south,yshift=8pt,xshift=0pt] at (current page.south) {\fbox{\parbox{\dimexpr\textwidth-\fboxsep-\fboxrule\relax}{\copyrighttext}}};
\end{tikzpicture}%
}

\makeatletter
\def\ps@IEEEtitlepagestyle{%
  \def\@oddfoot{\copyrightnotice}%
  \def\@evenfoot{}%
}

\begin{document}

\title{Fast Graph-Based Object Segmentation for RGB-D Images}

\author{Giorgio Toscana, Stefano Rosa\\
Politecnico di Torino\\
Corso D. Degli Abruzzi 24, 10100, Turin, Italy\\
{\tt\small name.surname@polito.it}
}

\maketitle

\begin{abstract}
Object segmentation is an important capability for robotic systems, in particular for grasping.
We present a graph-based approach for the segmentation of simple objects from RGB-D images. We are interested in segmenting objects with large variety in appearance, from lack of texture to strong textures, for the task of robotic grasping. The algorithm 
does not rely on image features or machine learning. We propose a modified Canny edge detector for extracting robust edges by using depth information and two simple cost functions for combining color and depth cues. The cost functions are used to build an undirected graph, which is partitioned using the concept of internal and external differences between graph regions. The partitioning is fast with ${\cal O}(N log N)$ complexity. We also discuss ways to deal with missing depth information.
We test the approach on different publicly available RGB-D object datasets,
such as the Rutgers APC RGB-D dataset and the RGB-D Object Dataset, and compare the results with other existing methods.
\end{abstract}

\IEEEpeerreviewmaketitle

\section{Introduction}
New and affordable depth sensors, like the Kinect, have been of great interest to the robotics community. These new sensors are able to simultaneously capture high-resolution color and depth images at high frame rates (RGB-D images).

Most recent work on RGB-D perception has been targeted to semantic segmentation or \emph{labelling} \cite{gupta2013perceptual,ren2012rgb}, namely the task of assigning a category label to each pixel of the image. 
While this is an important problem, many practical applications require a richer understanding of the scene. In particular, the notion of object instances is missing from such algorithms. 

Object detection in RGB-D images \cite{lai2011large, kim2013accurate}, in contrast, focusses on object instances, but the typical output of these approaches is a bounding box. 

Neither of these approaches produces a useful output representation for robot grasping \cite{hariharan2014simultaneous}. 
In some cases it may not be enough to know that an area of the image contain object pixels; at the same time, a bounding box of an individual object instance may not be precise enough for the robot to grasp it. The best approach for this kind of task is \emph{instance segmentation}, which consists in delineating all the object pixels corresponding to each detection.

For instance, in \cite{6163000} the authors propose a real-time algorithm for the segmentation of RGB-D sequences.  Segments represent the equilibrium states of a Potts model. Interaction strength between pixels  is based on color difference and is penalized based on large depth discrepancies. The Potts equilibrium is then found using a Metropolis algorithm implemented on GPU. The method however may not be robust to strong textures on objects or background, and thus produce over-segmentation.

In \cite{mishra2009active} a biologically-inspired algorithm was proposed for segmenting regions around a set of \emph{fixation points}. Monocular cues and depth or motion information (from stereo vision or optical flow) cues are combined in an independent way. 
In \cite{mishra2012segmenting} the approach was extended with a strategy for automatically selecting fixation points inside "simple" objects, by estimating if borders belong to the inner or outer side of objects. The algorithm does not explicitly deal with missing depth information in computing the border ownership.    
In \cite{rao2010grasping} the authors extend a graph-based segmentation algorithm \cite{felzenszwalb2004efficient} by integrating depth as a fourth component among RGB components in computing difference between adjacent pixels.

More recently, in \cite{gupta2014learning} semantically rich image and depth features have been used for object detection in RGB-D images, based on geocentric embedding for depth images that encodes height above ground and angle with gravity for each pixel in addition to horizontal disparity. Segmentation was also performed by labelling pixels belonging to object instances found by a Neural Network detector; decision forests were used for region classification. The approach uses GPU.

The contribution of this work is the extension of a fast graph-based segmentation algorithm\cite{felzenszwalb2004efficient} with the inclusion of depth information (similarly to \cite{rao2010grasping}) and saliency. 

We focus on the problem of object detection for robotic grasping, and in particular we are most interested in the Amazon Picking Challenge \cite{apc} scenario.
One aspect of the competition is to attempt simplified versions of the general task of picking items from shelves. The robots are presented with a stationary and lightly cluttered inventory shelf and are asked to pick a subset of the products and put them on a table.
We first propose a modified algorithm for depth image smoothing which takes into account depth shadows. We then describe a modified Canny edge detector that integrates depth information for finding robust edges. Then we propose two cost functions for creating a weighted graph that will be partitioned into regions. We finally use some rejection steps to discard most of the regions that do not belong to objects.

In Section \ref{sec:background} we introduce the problem of graph-based image segmentation; in Section \ref{sec:segmentation} we
present our approach; in Section \ref{sec:results} we present and discuss experimental results and finally in Section \ref{sec:conclusion}
we draw conclusions and illustrate future work.

\section{Background}
\label{sec:background}
In this work we started from the segmentation algorithm described in \cite{felzenszwalb2004efficient}, which is based on graph formalism. Let $\cal G = (\cal V,\cal E)$ be an undirected graph with vertices ${\cal V} = (v_i, \dots, v_{Np})$  corresponding to image pixels, and edges $e_{ij} = (i, j) \in \cal E$ , that connect pairs of neighboring vertices, namely $v_i$ and $v_j$. Each edge $e_{ij}$ has a corresponding weight $w_{ij}$, which is a measure of the similarity between $v_i$ and $v_j$. Hence image segmentation reduces to partitioning of $\cal G$ in subgraphs sharing similar characteristics.

Edge weights can be computed by evaluating color or intensity difference. If the function $\Lambda : \mathbb{R}^2 \rightarrow: \mathbb{R}^5$ associates each node (pixel) $v_j$ with the corresponding feature vector containing both node coordinates $v_{i_x},v_{i_y}$ and the  
RGB values $v_{i_r},v_{i_g},,v_{i_b}$, then the edges weights are computed as:
\beq
w_{ij}=\vert\vert{\Lambda(v_i)-\Lambda(v_j)} \vert\vert, \forall (i,j) \in {\cal E}.
\eeq
Let's define the \emph{internal difference} within the region $R_a$ as:
\beq
{\cal I}(R_a)= \max_{(i,j)\in {\cal E},\, {i,j} \in R_a} w_{ij},
\eeq
The segmentation procedure starts by considering each pixel as a different region. Such regions are pairwise compared and two regions are merged together in a bigger cluster if the following condition holds:
\beq
{\cal M}(R_a, R_b) \le \min ( {\cal I}(R_a)+ {\gamma \over \vert R_a \vert}, {\cal I}(R_b) + {\gamma \over {\vert R_b \vert}}),
\label{eq:merging}
\eeq
otherwise a boundary exist between them. In (\ref{eq:merging}), $\gamma$ is a constant parameter and the operator $\vert \cdot \vert$ returns the region size in pixels.

The described approach considers the internal characteristics of each region in the pairwise comparison, hence it is effective in segmenting image scenes with texture or non-uniform colors; moreover it is efficient, requiring a complexity of ${\cal O}(N_p log N_p)$, where $N_p$ is the number of pixels in the image.


\begin{figure}
\begin{tabular}{cc}
  \includegraphics[width=.45\columnwidth]{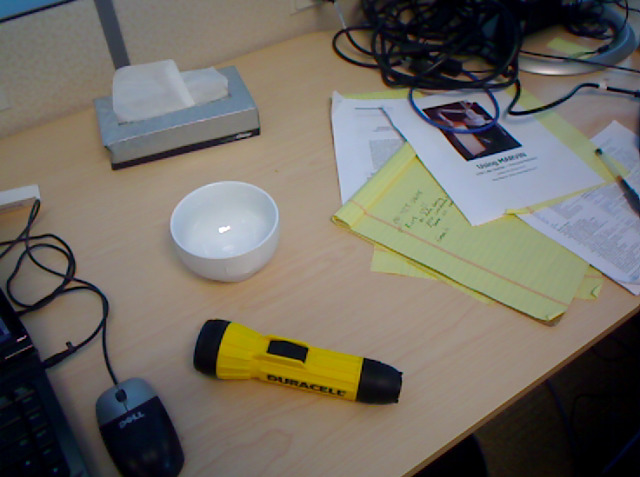} &   \includegraphics[width=.45\columnwidth]{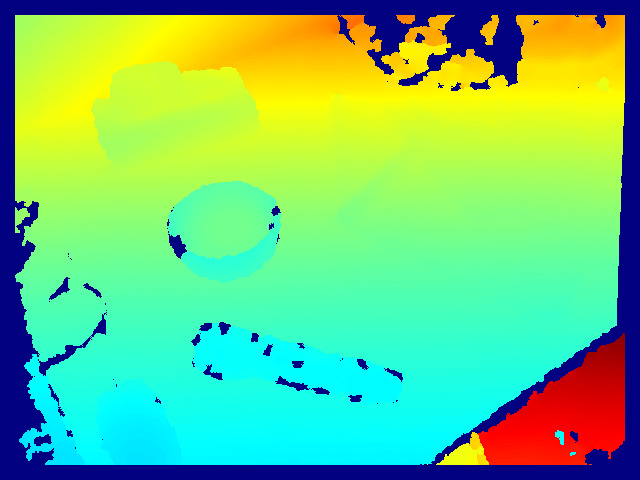} \\
(a) & (b) \\[6pt]
 \includegraphics[width=.45\columnwidth]{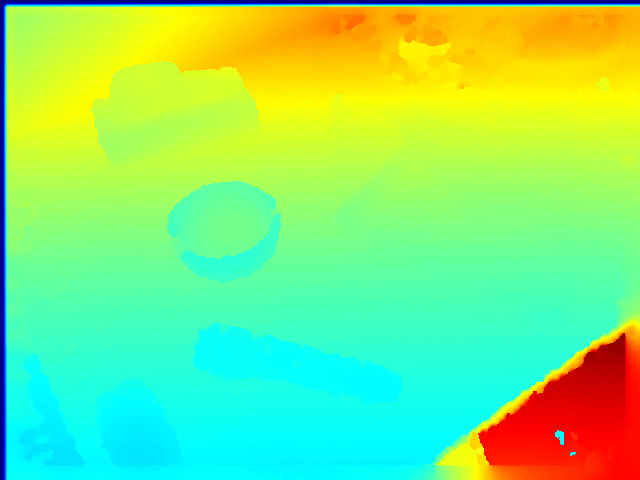} &   \includegraphics[width=.45\columnwidth]{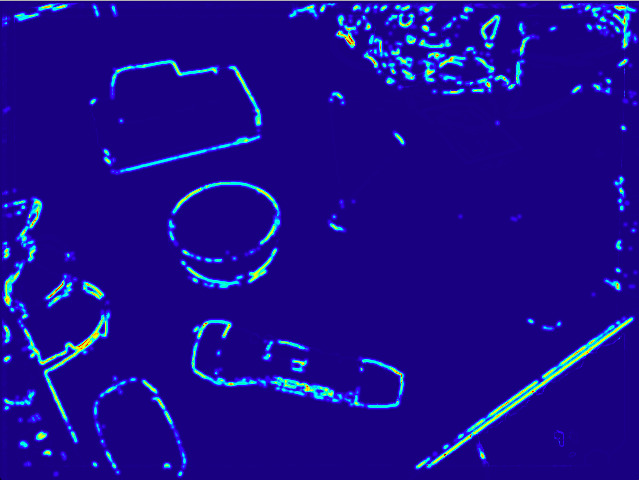} \\
(c) & (d) \\[6pt]
\includegraphics[width=.45\columnwidth]{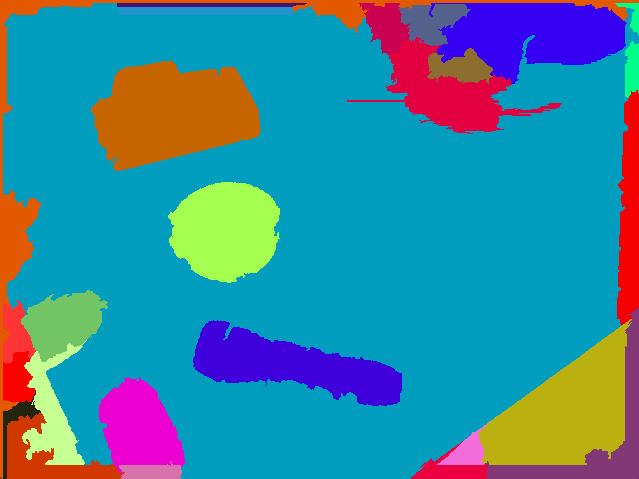} &
\includegraphics[width=.45\columnwidth]{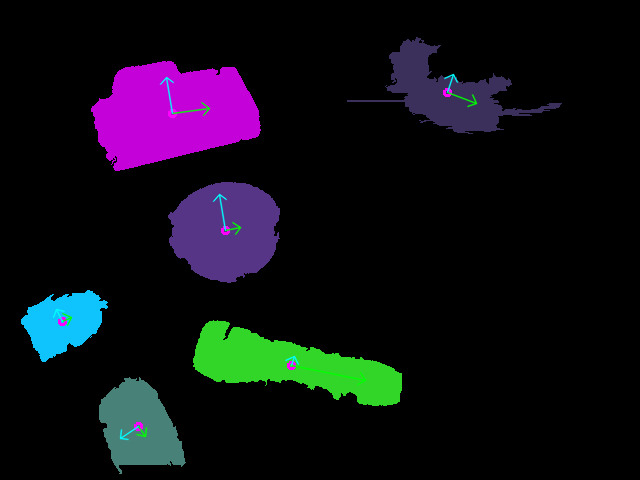} \\
(e) & (f) \\[6pt]
\multicolumn{2}{c}{\includegraphics[width=40mm]{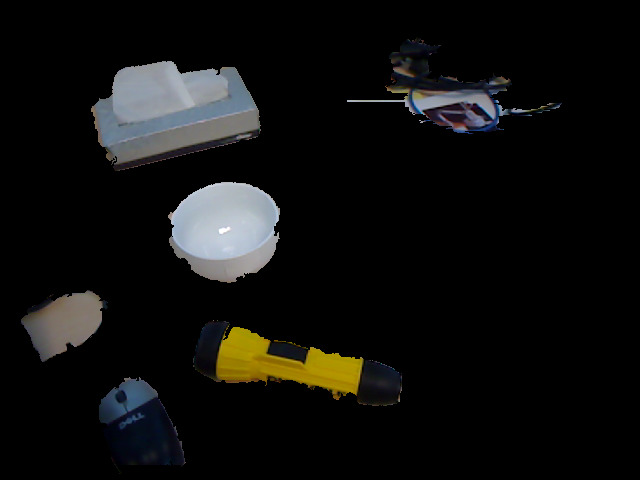} } \\
\multicolumn{2}{c}{(g)} \\[6pt]
\end{tabular}
\caption{The algorithm in action. (a) RGB image; (b) depth image; (c) smoothed and inpainted depth image; (d) obtained graph weights (color coded); (e) segmentation result; (f) result after post-processing; (g) segmented objects. }
\label{fig:algorithm}
\end{figure}

\section{Enhanched Segmentation}
\label{sec:segmentation}
In this section we describe the approach in detail.
An example of the algorithm in action is shown in Figure \ref{fig:algorithm}.

\subsection{Color}
The color difference is computed in the HSV color space and is derived from the metric described in \cite{6163000}. Separation of
color information (hue, saturation) and intensity (value) is more robust for objects with shadows or changes in lightness.
The color difference between two vertices $v_i=(v_{i_h},v_{i_s},v_{i_v})$ and $v_j=(v_{j_h},v_{j_s},v_{j_v})$ is defined as:
\beq\label{eq:hsv}
 \delta_{ij_{hsv}}= { \sqrt{\delta_v^2+\delta_s^2} \over \sqrt{k_{dv}^2+k_{ds}^2}} ,
\eeq
where:
$$ \delta_v = k_{dv} \vert v_{i_v}-v_{j_v} \vert,  \delta_h=\vert v_{i_h}-v_{j_h} \vert, $$
$$ \theta = \begin{cases} \delta_h, & \mbox{if } \delta_h <180^{\circ} \\ 360^{\circ}-\delta_h, & \mbox{if } \delta_h \ge 180^{\circ} \end{cases} $$
$$\delta_s=k_{ds}\sqrt{v_{i_s}^2+v_{j_s}^2- 2 v_{i_s} v_{j_s} \cos{\theta}}. $$
The denominator in (\ref{eq:hsv}) is introduced to normalize the color error in the range $\left[0,1\right]$. The parameters $k_{dv}$,$~k_{ds}$ are used to weight the \textit{value} and \textit{saturation} differences respectively. In our experiments these parameters are always kept fixed to $k_{dv}=4.5$,$~k_{ds}=0.1$.

\subsection{Depth}
Since depth maps obtained from low-cost Kinect-like sensors are usually noisy and prone to quantization errors,
we first apply a smoothing to the depth map. In \cite{rusu}, a depth-dependent smoothing algorithm is proposed which generates a smoothing kernel of different size for each pixel in the depth image. The area of such a kernel is based on two indicators, i.e. depth information of the pixel itself and the distance of the pixel from the object borders. The former is used to generate a wider smoothing area for pixels far from the camera, since the noise of the depth data is proportional to the distance from the sensor. 
The latter guarantees that the object edges are not smoothed by the filter.
The final smoothing kernel sizes are saved in a \emph{Smoothing Area Map} $\mathcal{S}(y,x)$. The average value within a region is thus computed by means of the depth integral image $\mathcal{I}_D(y,x)$ and saved in the smoothed depth map $\mathcal{D}_s(y,x)$ as follows:
\begin{small}
\begin{equation}\label{eq:rusuSmoothing}
\begin{split}
\mathcal{D}_s(y,x) = &\frac{1}{\left(2r+1\right)^2} \left[~\mathcal{I}_D(y+r,x+r) - \mathcal{I}_D(y+r,x-r) \right.\\ &\left.- \mathcal{I}_D(y-r,x+r) + \mathcal{I}_D(y-r,x-r)~\right]; 
\end{split}
\end{equation}
\end{small}
where: $r = \mathcal{S}(y,x)~$.
We noticed that the smoothing from \cite{rusu} introduces noise when the depth image contains shadows (depth image pixels with zero depth value); this is because (\ref{eq:rusuSmoothing}) is not able to discriminate between pixels having real depth information and pixels having no depth component.
We therefore generate a binary image $\mathcal{B}_D(y,x) $ from the depth map $\mathcal{D}(y,x)$ as follows:
\begin{equation}\label{eq:binaryDepth}
\mathcal{B}_D(y,x) = 
\begin{cases}
    0    & \quad \text{if } \mathcal{D}(y,x)\neq 0 \\
    1    & \quad \text{if } \mathcal{D}(y,x) = 0 \\
\end{cases}
\end{equation}
The binary depth map integral image $\mathcal{I}_B(y,x)$ is then used to count the number of pixels with no depth information ($\gamma_0$) inside the smoothing area of a given depth image pixel.
\begin{equation}\label{eq:numNaNSmoothing}
\begin{split}
\gamma_0 =  & \mathcal{I}_B(y+r,x+r) - \mathcal{I}_B(y+r,x-r)~+  \\
            & - \mathcal{I}_B(y-r,x+r) + \mathcal{I}_B(y-r,x-r);           
\end{split}
\end{equation}
where: $r = \mathcal{S}(y,x)~$.\\

\begin{figure}
\begin{tabular}{cc}
  \includegraphics[width=.45\columnwidth]{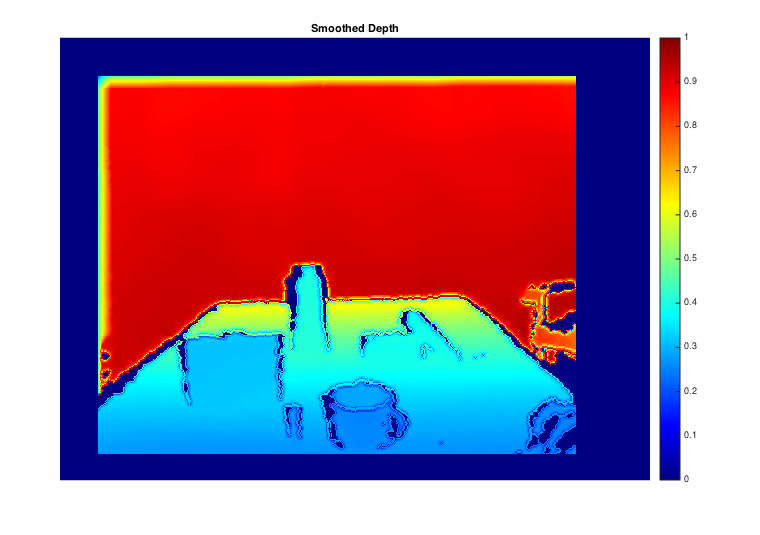} &   \includegraphics[width=.45\columnwidth]{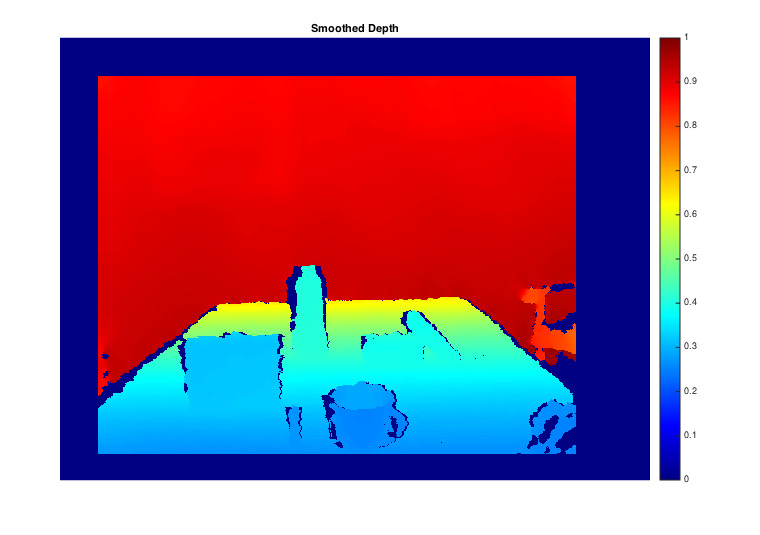} \\
(a) & (b) \\[2pt]
 \includegraphics[width=.45\columnwidth]{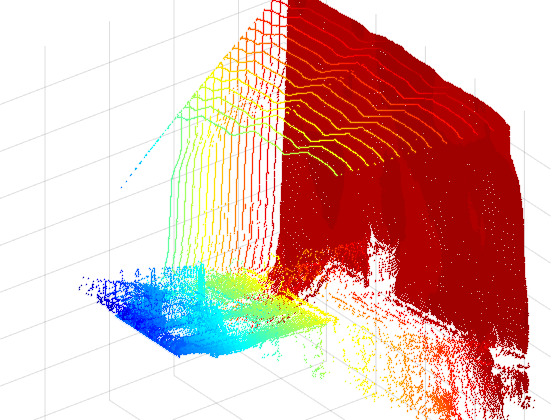} &   \includegraphics[width=.45\columnwidth]{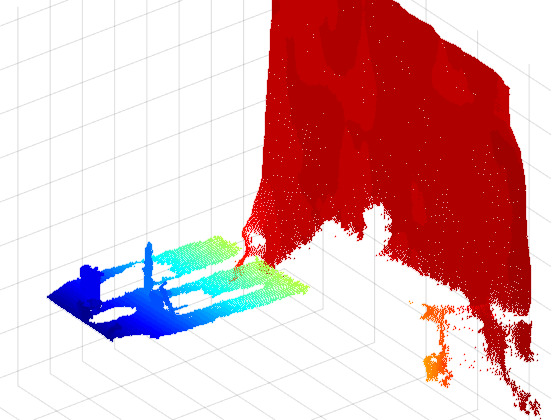} \\
(c) & (d) \\[2pt]
 \includegraphics[width=.45\columnwidth]{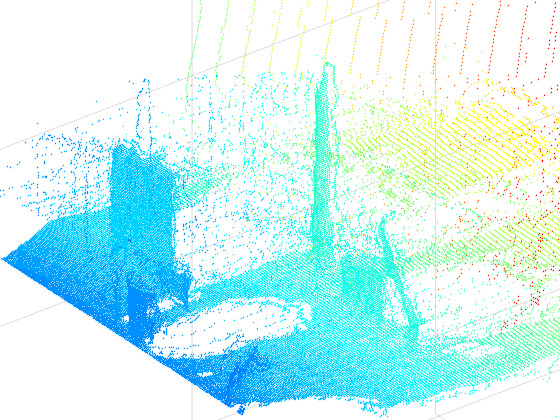} &   \includegraphics[width=.45\columnwidth]{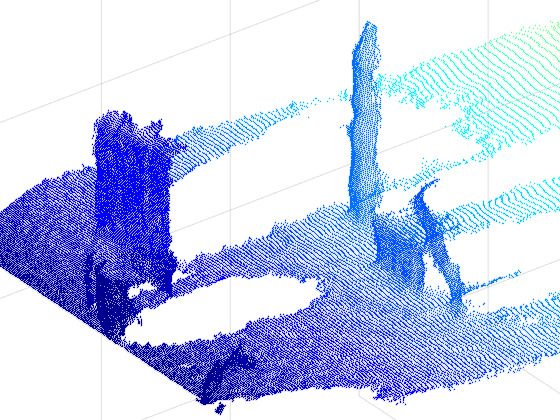} \\
(e) & (f) \\[2pt]

\end{tabular}
\caption{Comparison of depth smoothing algorithms in presence of missing depth values (dark blue pixels in both (a) and (b) images). Left column shows the results of the original smoothing algorithm as described in \cite{rusu}. Right column depicts the depth smoothing results obtained through our own changes to the original algorithm.  (a) and (b) show $\mathcal{D}_s(y,x)$. (c),(d) Show the PointCloud extracted from $\mathcal{D}_s(y,x)$. (e) and (f) show object and edge details.}
\label{fig:mySmoothing}
\end{figure}

(\ref{eq:rusuSmoothing}) is thus updated 
as shown in (\ref{eq:MySmoothing}).
\begin{small}
\begin{equation}\label{eq:MySmoothing}
\begin{split}
&\mathcal{D}_s(y,x) = \frac{1}{\left(2r+1\right)^2-\gamma_0} \left[~\mathcal{I}_D(y+r,x+r)~ + \right.\\ &\left. - \mathcal{I}_D(y+r,x-r) - \mathcal{I}_D(y-r,x+r) + \mathcal{I}_D(y-r,x-r)~\right];
\end{split}
\end{equation}
\end{small}
The denominator in (\ref{eq:MySmoothing}) equals zero if and only if all depth image pixels inside the smoothing kernel are equal to zero (no depth data is available).
In this case the smoothing has no meaning and $\mathcal{D}_s(y,x) = \mathcal{D}(y,x)$.\\
Figure \ref{fig:mySmoothing} shows a comparison between the original depth smoothing algorithm and the modified one.

The depth error between two vertices $v_{i_d}=\left(y_{i_d},x_{i_d}\right)$ and $v_{j_d}=\left(y_{j_d},x_{j_d}\right)$ is then defined as:
\beq
\delta_{ij_{depth}}=\mathcal{D}_s(y_{i_d},x_{i_d})-\mathcal{D}_s(y_{j_d},x_{j_d})
\eeq
$\delta_{ij_{depth}}$ is then normalized to be in the range $\left[0,1\right]~$.

When either $v_{i_d}$ or $v_{j_d}$ is undefined (one of the pixels belongs to a shadow in the depth image) $\delta_{ij_{depth}}$ is set to 0, since a shadow border is not necessarily an object border. Shadows in Kinect-like depth images can be caused by several effects: occlusions, highly reflective or absorbing materials, highly skewed surfaces, thin objects, or objects placed too near to the sensor.

\subsection{Saliency}
\begin{figure}
\begin{tabular}{cc}
  \includegraphics[width=.45\columnwidth]{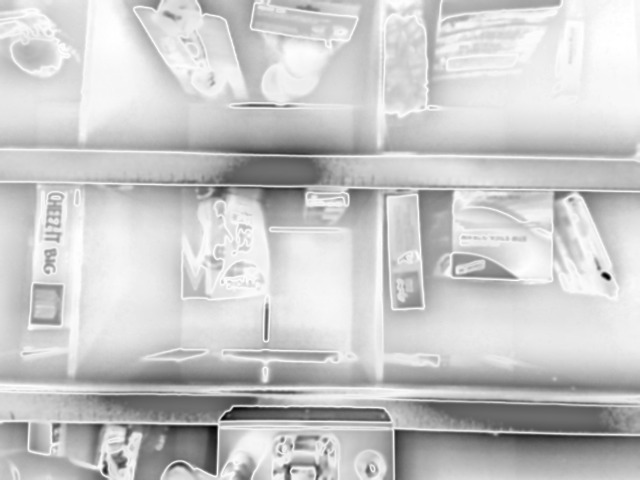} &   \includegraphics[width=.45\columnwidth]{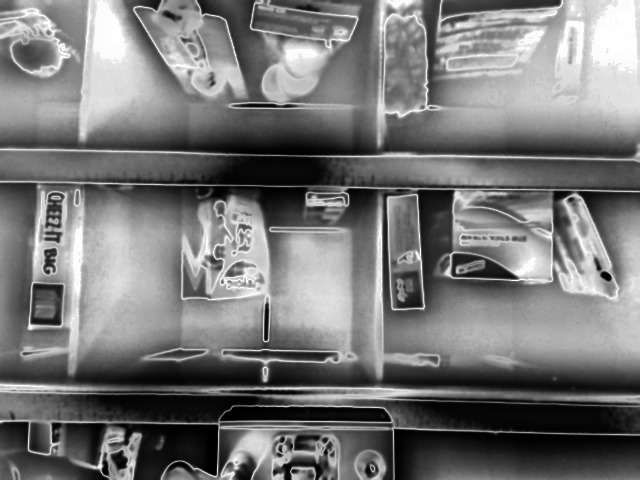} \\
(a) & (b) 
\end{tabular}
\caption{(a) Original saliency image. (b) Power-law filtered saliency image}
\label{fig:SaliencyPowerlaw}
\end{figure}

We compute a visual saliency map $\mathcal{V}_S(y,x)$ on the color image using the algorithm proposed in \cite{montabone2010human}, which is a fast implementation of visual saliency that uses an integral image on the original scale of the image in order to obtain high quality features in real time.
The algorithm is based on a single parameter for computing all the filter windows on a single integral image $\varsigma=\sigma 2^s$,
where $\sigma$ represents the surround and $s$ the scale.
The saliency map $\mathcal{V}_S(y,x)$ is first normalized in the range $\left[0,1\right]~$ and then filtered through a power-law transformation as:
\beq
\hat{\mathcal{V}}_S(y,x) = \mathcal{V}_S(y,x)^{4}
\eeq
This transformation lowers the values of middle gray-level pixels while keeping the high gray-level ones (pixels close to white) almost unchanged. The latter are pixels that are likely to belong to object borders. Figure \ref{fig:SaliencyPowerlaw} shows the transformed saliency image.

The saliency for each vertex $v_{i_s} = \left(y_{i_s},x_{i_s}\right)$ is thus defined as:
\beq
\delta_{i_{sal}}= \hat{\mathcal{V}}_S\left(y_{i_s},x_{i_s}\right)
\eeq

\subsection{Removing texture edges}
Similarly to \cite{schafer2013depth} and \cite{mishra2012segmenting}, we also want to remove edges caused by strong textures by calculating the difference in depth of two pixels on opposite sides of an edge pixel. While the first approach averages the depth gradient of an edge pixel within a small neighborhood of pixel 
along a connected edge
and the second approach uses a logistic function trained over examples for detecting depth boundary edges, we run a Canny edge detector on the image, based on the Scharr kernels to obtain a binary edge map $\mathcal{E}_E(y,x)$. We also extract the gradient directions $\Theta$ of the edge pixels and we discretize them to one of the possible angle (namely $0^{\circ},45^{\circ},90^{\circ},135^{\circ}$).\\
For each edge pixel $\vec{e}=(y_e,x_e) \in \mathcal{E}_E$, we sample two points, one along the positive edge gradient direction and the other along the negative one and we compute the depth gradients as:
\begin{equation}
\begin{split}
\rho_{+} = \mathcal{D}_s(\vec{e}) - \mathcal{D}_s(\vec{e}+\varepsilon_{\rho}\vec{n})\\
\rho_{-} = \mathcal{D}_s(\vec{e}) - \mathcal{D}_s(\vec{e}-\varepsilon_{\rho}\vec{n})
\end{split}
\end{equation}
where $\vec{n}$ is the edge normal vector and $\varepsilon_{\rho}$ indicate the pixel to pick along the edge gradient direction.\\
A \emph{depth boundary} map $\mathcal{E}_B(y,x)$ is the computed as follows:
\begin{equation}
\mathcal{E}_B(\vec{e}) = 
\begin{cases}
    0    & \quad \text{if } \rho_{+} <  t_{\rho} \wedge \rho_{-} <  t_{\rho}\\
    1    & \quad \text{otherwise} \\
\end{cases}
\end{equation}

\begin{figure}
\begin{tabular}{cc}
  \includegraphics[width=.45\columnwidth]{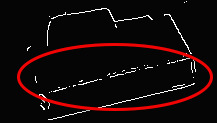} &   \includegraphics[width=.45\columnwidth]{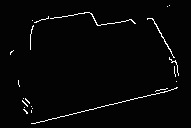} \\
(a) & (b)\\[6pt]
 \includegraphics[width=.45\columnwidth]{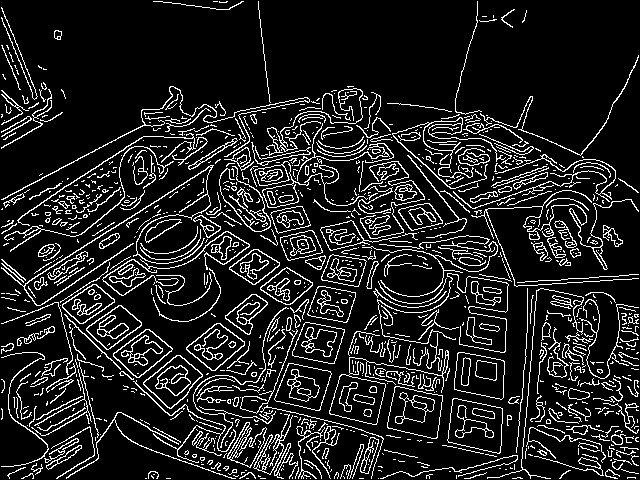} &   \includegraphics[width=.45\columnwidth]{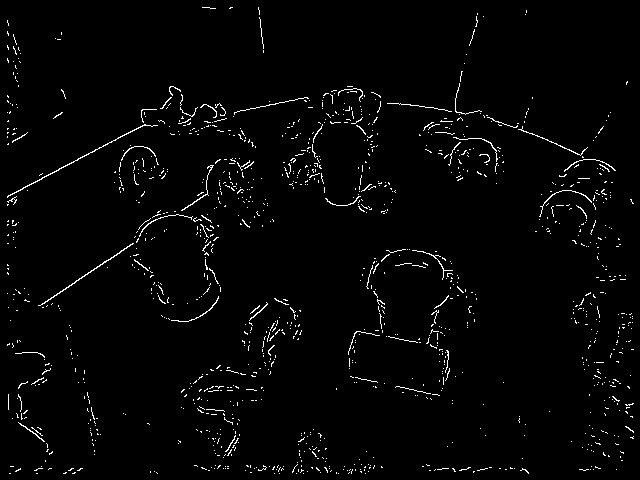} \\
(c) & (d) \\[6pt]
\includegraphics[width=.45\columnwidth]{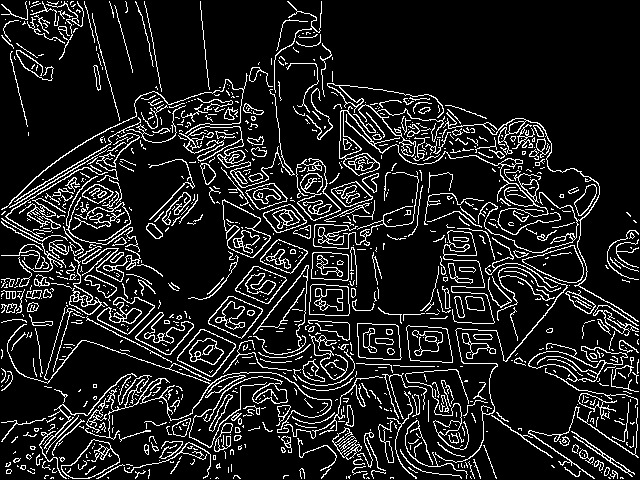} &   \includegraphics[width=.45\columnwidth]{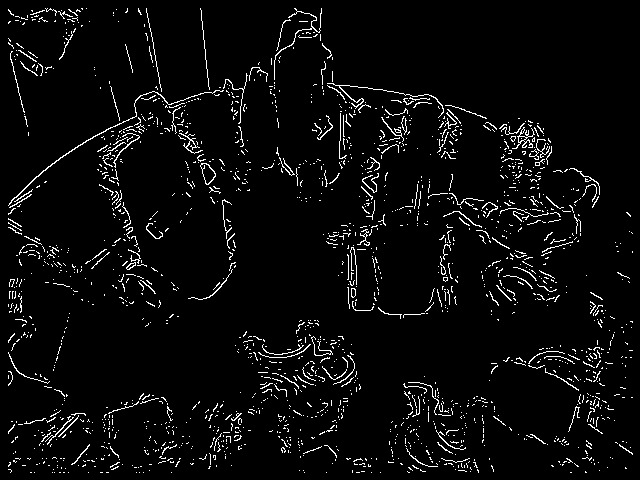} \\
(e) & (f)\\[6pt]
\multicolumn{2}{c}{\includegraphics[width=80mm]{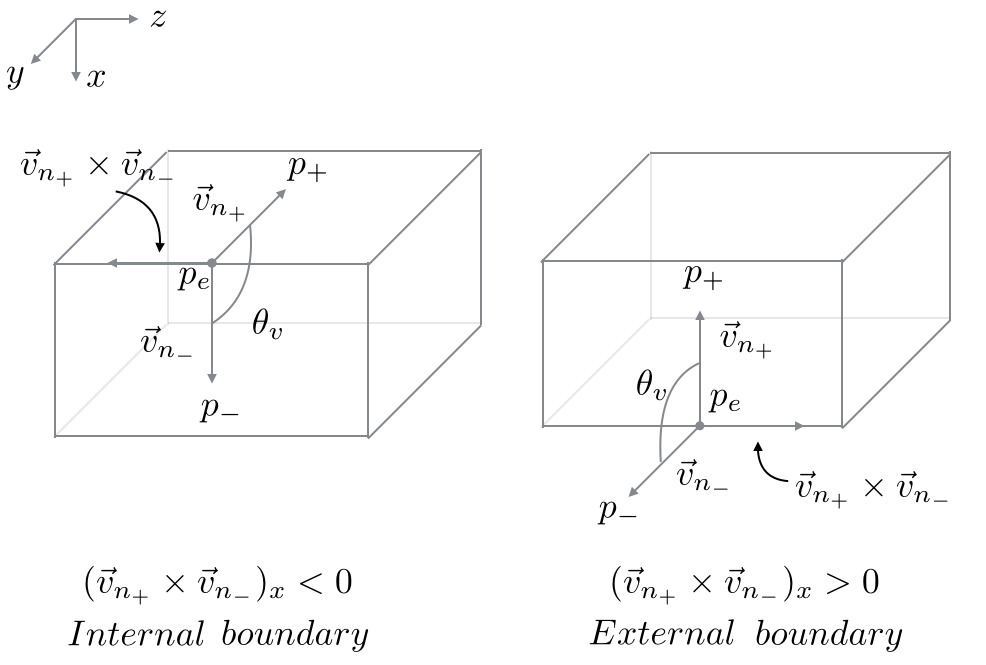} }\\
\multicolumn{2}{c}{(g)}
\end{tabular}
\caption{(a) Final Edge map with internal boundaries. (b) Final Edge map. (c) Canny output with no texture edge filter (Coffe Cups). (d) Coffe Cups Final Edge map. (e) Canny output with no texture edge filter (Milk Jugs). (f) Milk Jugs Final Edge map. (g) Internal boundary pixels definition.}
\label{fig:internalBoundary}
\end{figure}

If both $\rho_{+},~\rho_{-}$ are below a given threshold $t_{\rho}$, then the edge point $\vec{e}$ does not represent a real edge, but it is caused by texture instead; otherwise the point is a \emph{depth boundary}.
While most boundary pixels of an object correspond to depth discontinuities, the part of the object that touches the surface it is resting on does not have present depth discontinuity across it, therefore the \emph{contact edge} pixels are filtered out from the \emph{depth boundary} map $\mathcal{E}_B(y,x)$.\\
For each edge pixel $\vec{e}=(y_e,x_e) \in \mathcal{E}_E$, three points along the edge gradient direction are sampled, with the central one being the pixel $\vec{e}$. These three pixels are then projected onto the camera frame by means of the camera intrinsic matrix $\textbf{K}$ in order to obtain the corresponding three points: $\vec{p}_e,\vec{p}_+,\vec{p}_- \in \mathbb{R}^3$.
\begin{equation}
\begin{split}
&\vec{p}_e = \textbf{K}^{-1}\left[ ~\mathcal{D}_s(\vec{e})~\tilde{\textbf{e}} ~\right]
\\ 
&\vec{p}_+ = \textbf{K}^{-1}\left[ ~\mathcal{D}_s(\vec{e}_+)~\tilde{\textbf{e}}_+ ~\right]
\\
&\vec{p}_- = \textbf{K}^{-1}\left[ ~\mathcal{D}_s(\vec{e}_-)~\tilde{\textbf{e}}_- ~\right]
\end{split}
\end{equation}
where:
\begin{equation*}
\vec{e}_+ = \vec{e} + \varepsilon_e\vec{n}; \quad \vec{e}_- = \vec{e} - \varepsilon_e\vec{n}; \quad \tilde{\textbf{e}} = \left[x_e,y_e,1\right]^{T} .
\end{equation*}
The two unit vectors defined by the above three points are computed together with their angle as follows:
\begin{align}\label{eq:3pointAngle}
&\vec{v}_{n_+} = \dfrac{\vec{p}_+-\vec{p}_e}{\| \vec{p}_+-\vec{p}_e \|}\\
&\vec{v}_{n_-} = \dfrac{\vec{p}_--\vec{p}_e}{\| \vec{p}_--\vec{p}_e \|}\\
&\theta_v =  {\rm atan2}\left(\dfrac{\vec{v}_{\times}}{\vec{v}_{dot}}\right)\label{eq:contactAngle}
\end{align}
where: $\vec{v}_{\times} = \vec{v}_{n_+} \times \vec{v}_{n_-}$ and $\vec{v}_{dot} = \vec{v}_{n_+} \cdot \vec{v}_{n_-}$.
For everyday objects lying on ordinary surfaces (such as tables, shelves, floors, etc.),  contact edge pixels can be estimated straightforward by filtering the angle $\theta_v$ in (\ref{eq:contactAngle}). Common interactions between objects and holding surfaces lead to contact angles close to $90^{\circ}$.\\
A \emph{contact boundary} map $\mathcal{E}_C(y,x)$ is the computed as follows:
\begin{equation}\label{eq:contactBoundary}
\mathcal{E}_C(\vec{e}) = 
\begin{cases}
    0    & \quad \text{if } \theta_v >  t_{\theta_H} \vee~ \theta_v <  t_{\theta_L}\\
    1    & \quad \text{otherwise} \\
\end{cases}
\end{equation}
where $t_{\theta_H}$ and $t_{\theta_L}$ are high and low thresholds used to cope with contact angles perturbations around the ideal value.\\
The contact boundary map as defined by (\ref{eq:contactBoundary}) and (\ref{eq:contactAngle}). It also includes false contact edges called \emph{internal boundaries} as shown in figure \ref{fig:internalBoundary}a. Since we are only interested in the objects external contours (e.g., figure \ref{fig:internalBoundary}b), we cancel internal edge pixels out by looking at the direction of the cross product $\vec{v}_{\times}$ as depicted in figure (\ref{fig:internalBoundary}g). 
Note that the vectors defined until now are all expressed w.r.t the camera reference frame with the $Z$-axis pointing outwards along the optical axis and the $X$-axis pointing to the right.
A contact edge pixel $\vec{i} = (y_i,x_i) \in \mathcal{E}_C(y,x)$ is estimated to be an internal boundary pixel if and only if the $x$ component of the cross product $\vec{v}_{\times}$ is less than zero. (\ref{eq:contactAngle}) is then updated as follows:
\begin{equation}\label{eq:contactAngleMod}
\theta_v = 
\begin{cases}
    - {\rm atan2}\left(\dfrac{\vec{v}_{\times}}{\vec{v}_{dot}}\right)    & \quad \text{if } v_{\times_x} < 0\\
    {\rm atan2}\left(\dfrac{\vec{v}_{\times}}{\vec{v}_{dot}}\right)    & \quad \text{otherwise} \\
\end{cases}
\end{equation}
The \emph{contact boundary} map $\mathcal{E}_C(y,x)$ definition (see \ref{eq:contactBoundary}) is therefore left unchanged.\\
The \emph{final boundary} map $\mathcal{E}_F(y,x)$ is then computed as:
\begin{equation}
\mathcal{E}_F(y,x) = \mathcal{E}_B(y,x) \cup \mathcal{E}_C(y,x)
\end{equation}
Despite the fact that the simple texture edge filtering might be seen as limited to a small class of simple objects (i.e., prismatic ones), Figures \ref{fig:internalBoundary}d and \ref{fig:internalBoundary}f show how the algorithm is very effective for other classes of objects  as well (i.e., cylindrical ones) and, in general, is able to handle complex object shapes (e.g., milk jugs).



The boundary for each vertex $v_{i_b} = \left(y_{i_b},x_{i_b}\right)$ is thus defined as:
\beq
\delta_{i_{bound}}= \mathcal{E}_F(y_{i_b},x_{i_b})
\eeq

\subsection{Weight function}
Two cost functions are proposed. The first one includes color, depth and saliency information and it is found to work best when a large number of shadows are present in the depth image. The second one includes depth, color and boundary edges and works better when full depth information is available.

In the first cost function, the difference between two vertices $v_i$ and $v_j$ is defined as:
\beq\label{eq:w1}
\small
\begin{aligned}
w_{ij}=
{
		{k_y \log_{2}(1+\delta_{ij_{hsv}}) +  k_x \log_{2}(1+\delta_{ij_{depth}})}
		\over
		{2+k_x+k_y+k_s}
} \\
+
{
	{
		\delta_{ij_{depth}} \delta_{i_{sal}}^{1+\delta_{ij_{depth}}} + \delta_{ij_{depth}} \delta_{ij_{hsv}}^{1+\delta_{ij_{depth}}}
		+ k_s \log_{2}(1+\delta_{i_{sal}})
	}
	\over
	{2+k_x+k_y+k_s}
},
\end{aligned}
\eeq
where $k_s$, $k_y$ , $k_x$ are parameters for weighting in the saliency map, the color and depth difference respectively.

In the second cost function,  the difference between two vertices  is defined as:
\beq\label{eq:w2}
w_{ij}= {{k_x \delta_{ij_{depth}} \log_{2}(1+ \delta_{ij_{hsv}}) + k_b\delta_{i_{bound}}} \over {k_x+k_b}},
\eeq
where $k_b$ is a parameter for weighting in boundary edges, while the denominator in both (\ref{eq:w1}) and (\ref{eq:w2}) is needed to normalize the weights between $\left[0,1\right]$.\\
We use base-2 logarithms since all the cues used as input to the weight functions ($\delta$) are in the range $\left[0,1\right]$; the dynamic range of the cues is thus left unchanged. Moreover, logarithmic functions map a narrow range of low cue values into a wider range of output levels while compressing higher values. This property tends to create edge weights that are spread within their own dynamic range rather than generating quasi-binary weights maps.\\
(\ref{eq:w1}) is composed by a logarithmic term which handles a single independent variable (i.e., $k\log_{2}(1+\delta)$) and a coupled term which relates two variables (i.e., $\delta_{depth} \delta_{hsv}^{1+\delta_{depth}}$). The former controls the effect of each input information independently through the parameters $k$. It plays a fundamental role when no depth data are available, since the coupled terms equal zero.
The latter, instead, biases the weights assignment by introducing depth information. The lower the depth variation is, the lower is the contribute of color and saliency cues; this happens in highly textured objects where two pixels, belonging to the same object surface, generate small depth gradient but large color (or saliency) difference. For large depth gradients the weight shows an exponential trend and reaches maximum together with the color (or saliency) difference.
The exponential shape of $w$ for medium-high depth gradients is needed to mitigate the linear contribute of $\delta_{depth}$. This situation may arise in presence of concave objects (i.e., ceramic bowls or horseshoe-like objects) where medium-high variations of the depth gradient do not necessarily mean that the corresponding two graph vertices belong to different objects. 
In this case, in fact, the function generates lower edge weights in presence of low and medium visual cue differences.

The cost function in (\ref{eq:w2}) is used when little or no shadows are present in the depth map. We fill the small gaps in the depth image by in-painting. Depth maps with large shadows are not in-painted since the reconstruction error would generate noise and false object boundaries. In this case, the depth map is not modified and the cost function in (\ref{eq:w1}) is used instead. 
The second weight function has a coupled term that trades the information of $\delta_{hsv}$ and $\delta_{depth}$ like the first weight function but without any saliency data. The idea of this terms is the same defined in (\ref{eq:w1}) but we noticed how the logarithmic term here works better than the exponential one. The second term adds a bias term $k_b$ when the vertex $v_i$ is a boundary pixel $p_i=(y_i,x_i) \in \mathcal{E}_F$.\\

The graph is partitioned using Disjoint-set Forests. At the first iteration each node represents a distinct region $R_i$. Regions are iteratively merged based on (\ref{eq:merging}).
The final result is a set of regions $\cal R$.


\subsection{Post-processing} 
In order to discard false positives, such as regions that belong to the background, some rejections steps are 
required on the set $\cal R$.

Principal component analysis is performed on each region to estimate the principal components $\vec{x_1}$, $\vec{x_2}$, the relative eigenvalues $\lambda_{1}$, $\lambda_{2}$ and its eccentricity $\varepsilon$.
If either $\lambda_{1}$, $\lambda_{2}$ or $\varepsilon$ are over given thresholds, the region is discarded. The threshold can be roughly estimated if the classes of objects to be found are known in advance.

We add two more rejection steps when dealing with difficult lightning conditions and poor depth maps (see Section \ref{sec:rutgers}).
When not using in-painting of the depth image, 
regions 
whose pixels with no valid depth data are greater than 30$\%$ of total region size are also discarded, as this may lead to the failure of the robot grasping policies defined thereafter. Finally, dark regions can be discarded too. A $32$ bins histogram of the brightness component of the region is computed. If $30\%$ of region pixels fall within the first three bins of the histogram (i.e., pixels values in the range $\left[0,24\right]$), the region is discarded.
Since we are interested in grasping, it is also possible to discard regions which are out of reach for a robotic arm, being too distant to the camera frame.


\section{Experimental results}
\label{sec:results}
We tested our approach on three public datasets of RGB-D scenes \cite{rutgers}, \cite{rgbd} and \cite{tejani2014latent}.
We also compare the results with the one proposed in \cite{mishra2012segmenting} and show a qualitative comparison with the original algorithm \cite{felzenszwalb2004efficient} and \cite{6163000}.
For each image of every dataset, objects have been manually labelled by delineating the pixels inside the object boundary. 
If the segmented objects overlap more than 70\% with the corresponding object pixels, we consider the object as successfully segmented, as in \cite{rao2010grasping}.

\begin{figure}
\begin{tabular}{cc}
 \includegraphics[width=.45\columnwidth]{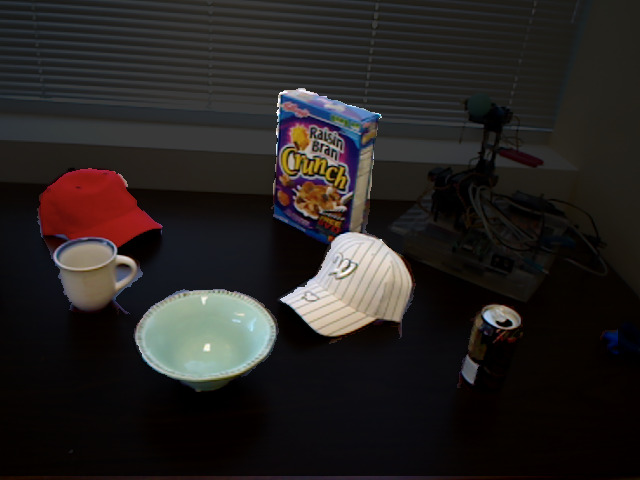} &   \includegraphics[width=.45\columnwidth]{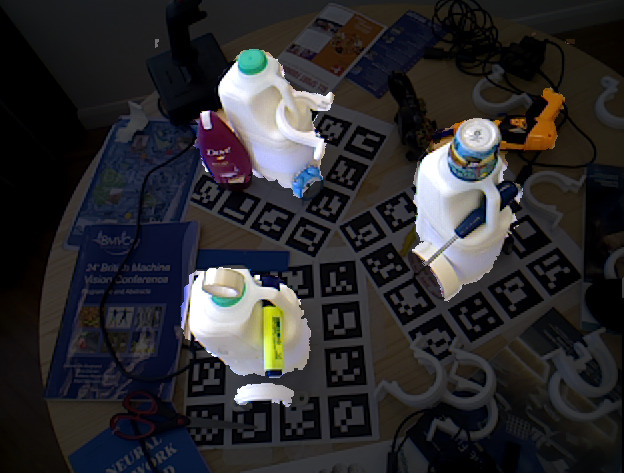} \\
(a) & (b) \\[6pt]
 \end{tabular} 
\caption{Examples of the results on the \cite{rutgers} dataset (a) and   \cite{tejani2014latent} dataset (b). Segmented objects are highlighted.}
\label{fig:challenge}
\end{figure}

\begin{figure}[h!]
\center
\begin{tabular}{cc}

          \includegraphics[width=.37\columnwidth]{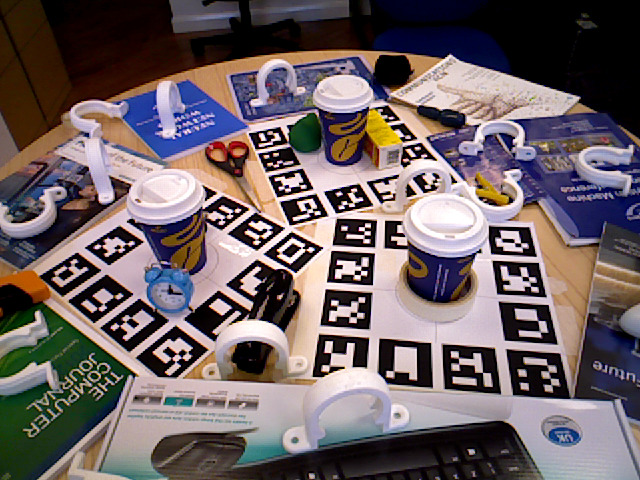} &
          \includegraphics[width=.37\columnwidth]{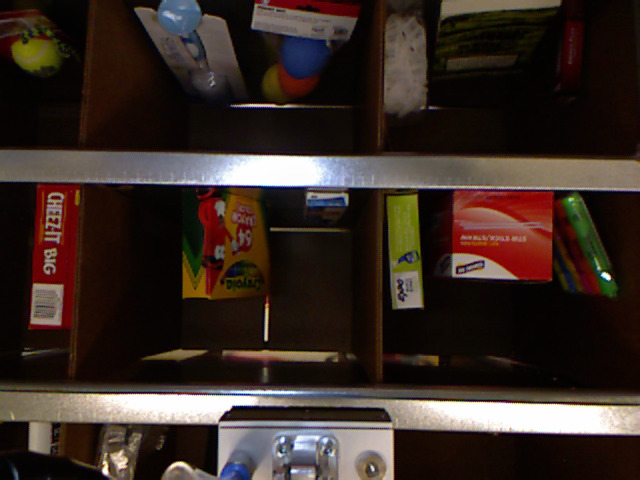}\\
          	\includegraphics[width=.37\columnwidth]{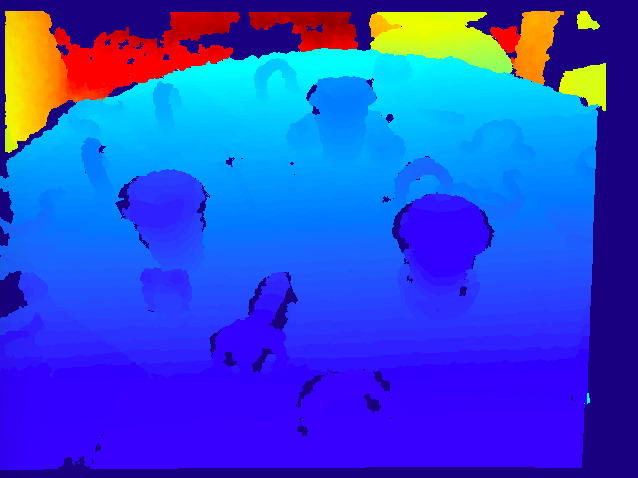} &
         	 \includegraphics[width=.37\columnwidth]{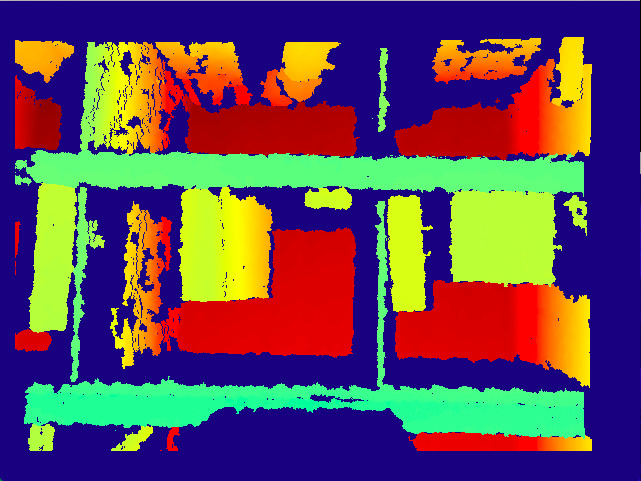}\\      
    	  \includegraphics[width=.37\columnwidth]{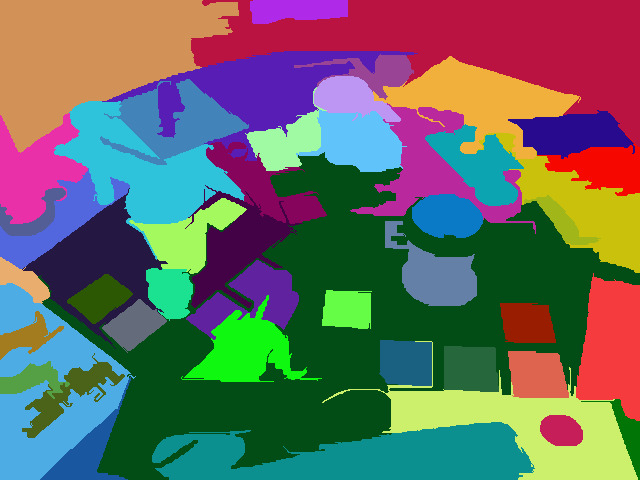} &
	  \includegraphics[width=.37\columnwidth]{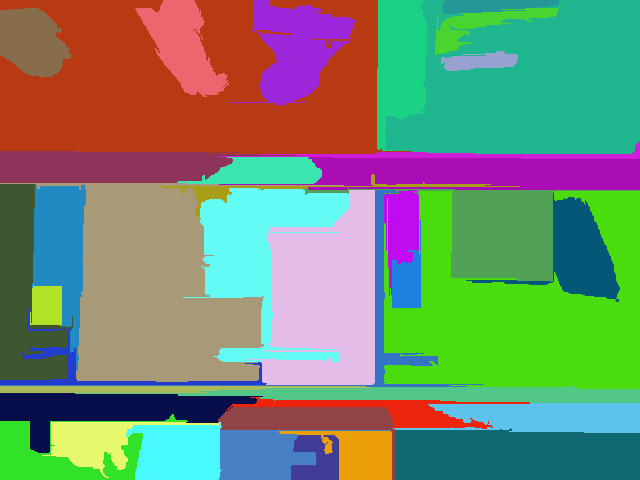}\\
          \includegraphics[width=.37\columnwidth]{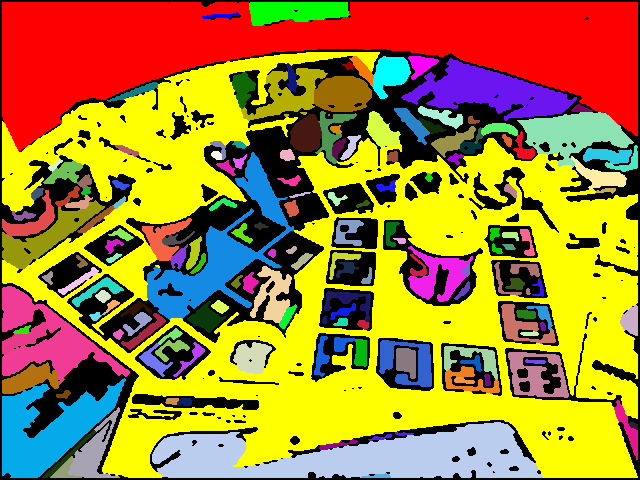} &
	  \includegraphics[width=.37\columnwidth]{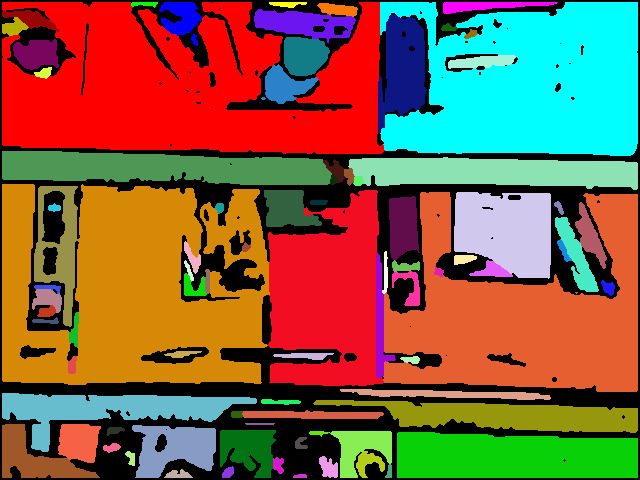}\\
	
	    	  \includegraphics[width=.37\columnwidth]{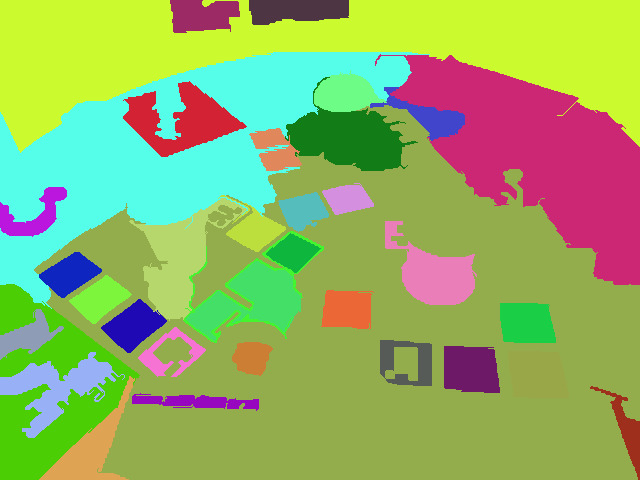} &
	  \includegraphics[width=.37\columnwidth]{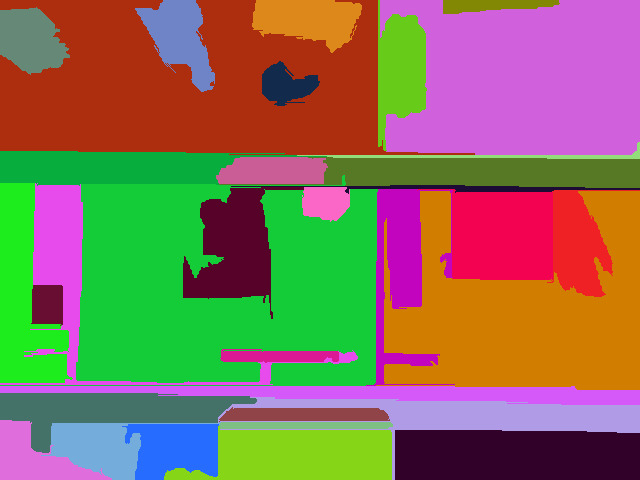}\\

          \includegraphics[width=.37\columnwidth]{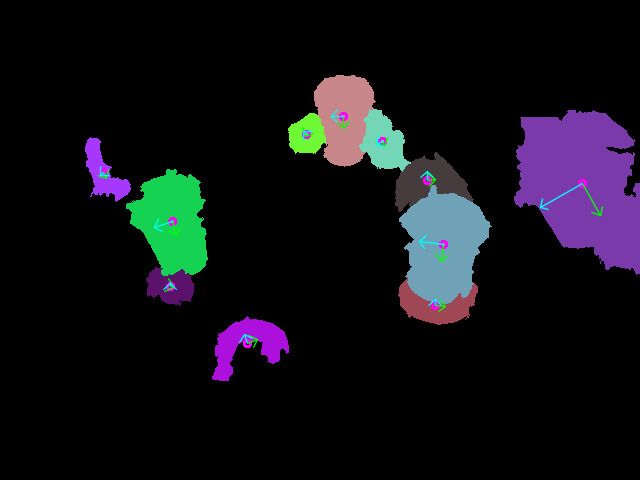} &
          \includegraphics[width=.37\columnwidth]{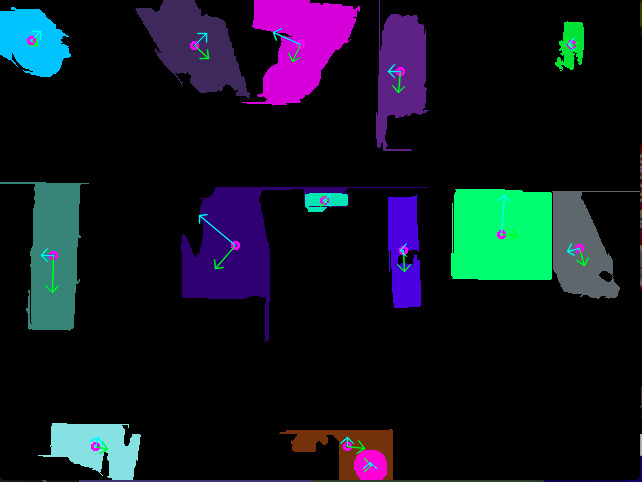}\\		
          \includegraphics[width=.37\columnwidth]{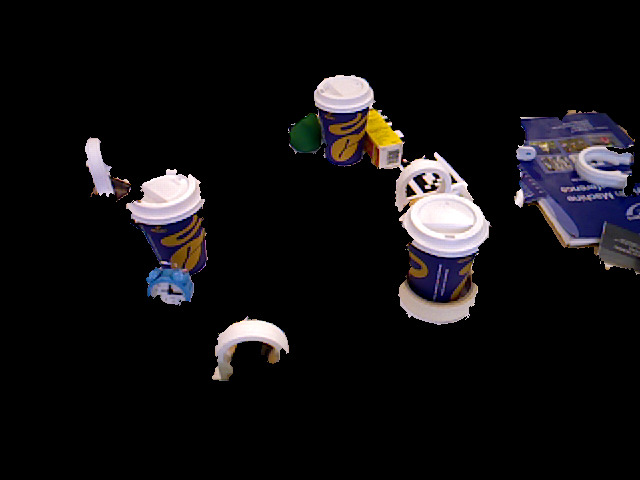} &
          \includegraphics[width=.37\columnwidth]{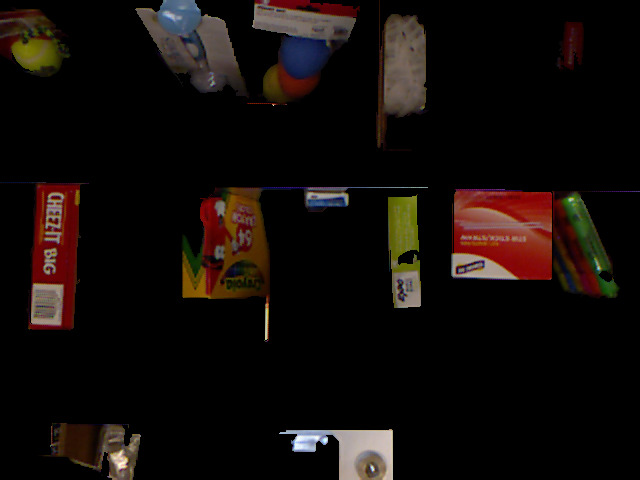}\\

 \end{tabular} 
       \caption{Comparison between different approaches. First and second row: original images; third row: \cite{felzenszwalb2004efficient}; fourth row: \cite{6163000}; fifth row: \cite{rao2010grasping}; sixth and seventh row: our approach.}
\label{fig:comparison}
\end{figure}

The software has been developed using the OpenCV library in C++ under Linux and runs on CPU. The source code will be available online. 
All frames are $640 \times 480$ and the average processing time per image was 0.6 s on a standard PC with a 2.3Ghz CPU (single thread). Figure \ref{fig:challenge} shows two results on different datasets, while Figure \ref{fig:comparison} shows a comparison between different approaches.

\subsection{Rutgers APC RGB-D Dataset}
\label{sec:rutgers}
This dataset has been created specifically for the Amazon Picking Challenge and is composed of different runs,
each one containing a series of RGB images and corresponding depth images, acquired using a Asus XTion sensor in 
different positions and with a variable number of objects on shelves.
For each position, four consecutive images are provided.

The dataset is particularly challenging due to low lightning and heavy presence of shadows and missing areas in the depth image. We assembled three runs from the dataset with increasing average number of objects in each image. 
We tested our approach on a subset of runs. We used the first weight function, due to the large number of shadows. 
Parameters have following values: $\gamma=5$, $k_x=1.05$, $k_y=1.5$, $k_s=0.5$. 
Results are reported in Table \ref{table:rutgers}.

\begin{table}
\begin{center}
\begin{tabular}{|l|c|c|}
\hline
 		& No. of objects & $\%$ of objects detected \\
\hline\hline
Run\_1  & 80   & 87.9\%  \\
Run\_2  & 120 & 92.3\% \\
Run\_3  & 121 & 75.6\%  \\
\hline
\end{tabular}
\end{center}
\caption{Results for the Rutgers APC Dataset. 
}
\label{table:rutgers}
\end{table}

\subsection{RGB-D Scenes Dataset}
The RGB-D Scenes Dataset consists of 8 scenes annotated with objects that belong to the RGB-D Object Dataset. (bowls, caps, cereal boxes, coffee mugs, and soda cans).  Each scene is a single video sequence consisting of multiple RGB-D frames. The objects are visible from different viewpoints and distances and may be partially or completely occluded.
We compare the results of the proposed algorithm with the one from \cite{mishra2012segmenting}. We tested the approach on subset of six objects and on three different scenes. 
We used the second weight function and set the parameters to the following values: $\gamma=0.0016$, $k_x=7.5$, $k_b=0.66$.
Results are shown in Table \ref{table:rgbd}. It should be noted that our approach does not rely on knowledge of the camera pose and is thus more general, at the cost of lower accuracy for some objects, while attaining 100\% accuracy for other objects. 
Results are comparable to \cite{mishra2012segmenting}, though the metric we use is more strict (in \cite{mishra2012segmenting} an overlap of 50\% is considered as a good detection).


\begin{table*}
\begin{center}
\begin{tabular}{|c|c|c|c|c|c|c|}
\hline
		& \multicolumn{6}{|c|}{$\%$ of objects detected} \\

		& Soda can & Coffee mug & Cap & Bowl & Flashlight & Cereal box \\
\hline\hline
Table\_1  & 90.6\% (100\%) & 100\% (83.6\%) & 80.1\% (93.6\%) & 85.5\% (90.3\%) & 98.1\% (98.1\%) & 72\% (97.8\%) \\

Desk\_1   & 100\% (93.7\%) & 100\% (92.5\%) & 74.2\% (100\%) & - & - & - \\
Kitchen\_small\_1 & 98.6\% (74.8\%) & 100\% (70.1\%) & 86.5\% (97.3\%) & 100\% (90\%) & 100\% (88.5\%) & 77.6\% (84.4\%) \\
\hline
\end{tabular}
\end{center}
\caption{Results for the RGB-D Dataset. Inside parentheses, the results from \cite{mishra2012segmenting} are reported for comparison.}
\label{table:rgbd}
\end{table*}

\subsection{Multiple-instance dataset}
In \cite{tejani2014latent}, 6 objects are captured under varying viewpoint with lots of background clutter, scale and pose changes, and in particular foreground occlusions and multi-instance representation 
(three instances of the same object are present in each frame as well as other objects and clutter). We tested the approach on a subset of scenes.
Parameters for the second weight function are as follows: $\gamma=0.001$, $k_x=1.2$, $k_b=0.05$.
Results are shown in Table \ref{table:challenge}. 

\begin{table}
\begin{center}
\begin{tabular}{|l|c|c|}
\hline
 		& No. of objects& $\%$ of objects detected \\
\hline\hline
Milk & 2589 & 66.6\% \\
Coffee\_Cup & 2127 & 87.2\% \\
Shampoo & 2118 & 99.6\% \\
Camera & 894 & 96.3\% \\

\hline
\end{tabular}
\end{center}
\caption{Results for the multiple instance dataset.}
\label{table:challenge}
\end{table}

\section{Conclusion}
\label{sec:conclusion}
We presented a fast approach for segmenting simple objects from RGB-D images in the 2D domain, without need for 3D models of the objects. The approach builds on graph-based image segmentation by integrating depth information and between object boundaries and strong texture and is able to work with texture-less objects as well as heavy textured ones. 
While the approach is targeted to robot grasping, it is general in its formulation. We specifically addressed the case of poor depth images and light conditions.
We proposed a modified Canny edge detector that integrates depth information in order to find robust edges in RGB-D images. We then proposed two cost functions for computing the graph weights.
The algorithm 
has been tested on three public object recognition datasets, relevant to robot grasping and containing different scenarios and challenges.
Future work will be devoted to parallelization of the graph creation and partitioning phases and to investigate different partitioning strategies. 
   
%
%


\bibliographystyle{plainnat}
\bibliography{full_biblio}

\begin{thebibliography}{18}
\providecommand{\natexlab}[1]{#1}
\providecommand{\url}[1]{\texttt{#1}}
\expandafter\ifx\csname urlstyle\endcsname\relax
  \providecommand{\doi}[1]{doi: #1}\else
  \providecommand{\doi}{doi: \begingroup \urlstyle{rm}\Url}\fi

\bibitem[apc()]{apc}
Amazon picking challenge.
\newblock Website.
\newblock \url{http://amazonpickingchallenge.org}.

\bibitem[rgb()]{rgbd}
Rgb-d object dataset.
\newblock Website.
\newblock \url{http://rgbd-dataset.cs.washington.edu}.

\bibitem[rut()]{rutgers}
Rutgers apc rgb-d dataset.
\newblock Website.
\newblock \url{http://http://pracsyslab.org/rutgers_apc_rgbd_dataset}.

\bibitem[Abramov et~al.(2012)Abramov, Pauwels, Papon, Worgotter, and
  Dellen]{6163000}
A.~Abramov, K.~Pauwels, J.~Papon, F.~Worgotter, and B.~Dellen.
\newblock Depth-supported real-time video segmentation with the kinect.
\newblock In \emph{Applications of Computer Vision (WACV), 2012 IEEE Workshop
  on}, pages 457--464, Jan 2012.
\newblock \doi{10.1109/WACV.2012.6163000}.

\bibitem[Felzenszwalb and Huttenlocher(2004)]{felzenszwalb2004efficient}
Pedro~F Felzenszwalb and Daniel~P Huttenlocher.
\newblock Efficient graph-based image segmentation.
\newblock \emph{International Journal of Computer Vision}, 59\penalty0
  (2):\penalty0 167--181, 2004.

\bibitem[Gupta et~al.(2014)Gupta, Girshick, Arbel{\'a}ez, and
  Malik]{gupta2014learning}
Saurabh Gupta, Ross Girshick, Pablo Arbel{\'a}ez, and Jitendra Malik.
\newblock Learning rich features from rgb-d images for object detection and
  segmentation.
\newblock In \emph{Computer Vision--ECCV 2014}, pages 345--360. Springer, 2014.

\bibitem[Gupta et~al.(2013)Gupta, Arbelaez, and Malik]{gupta2013perceptual}
Swastik Gupta, Pablo Arbelaez, and Jagannath Malik.
\newblock Perceptual organization and recognition of indoor scenes from rgb-d
  images.
\newblock In \emph{Computer Vision and Pattern Recognition (CVPR), 2013 IEEE
  Conference on}, pages 564--571. IEEE, 2013.

\bibitem[Hariharan et~al.(2014)Hariharan, Arbel{\'a}ez, Girshick, and
  Malik]{hariharan2014simultaneous}
Bharath Hariharan, Pablo Arbel{\'a}ez, Ross Girshick, and Jitendra Malik.
\newblock Simultaneous detection and segmentation.
\newblock In \emph{Computer Vision--ECCV 2014}, pages 297--312. Springer, 2014.

\bibitem[Holzer et~al.(2012)Holzer, Rusu, Dixon, Gedikli, and Navab]{rusu}
S.~Holzer, R.B. Rusu, M.~Dixon, S.~Gedikli, and N.~Navab.
\newblock Adaptive neighborhood selection for real-time surface normal
  estimation from organized point cloud data using integral images.
\newblock In \emph{Intelligent Robots and Systems (IROS), 2012 IEEE/RSJ
  International Conference on}, pages 2684--2689, Oct 2012.
\newblock \doi{10.1109/IROS.2012.6385999}.

\bibitem[Kim et~al.(2013)Kim, Xu, and Savarese]{kim2013accurate}
Byung-soo Kim, Shili Xu, and Silvio Savarese.
\newblock Accurate localization of 3d objects from rgb-d data using
  segmentation hypotheses.
\newblock In \emph{Computer Vision and Pattern Recognition (CVPR), 2013 IEEE
  Conference on}, pages 3182--3189. IEEE, 2013.

\bibitem[Lai et~al.(2011)Lai, Bo, Ren, and Fox]{lai2011large}
Kevin Lai, Liefeng Bo, Xiaofeng Ren, and Dieter Fox.
\newblock A large-scale hierarchical multi-view rgb-d object dataset.
\newblock In \emph{Robotics and Automation (ICRA), 2011 IEEE International
  Conference on}, pages 1817--1824. IEEE, 2011.

\bibitem[Mishra et~al.(2009)Mishra, Aloimonos, and Fah]{mishra2009active}
Ajay Mishra, Yiannis Aloimonos, and Cheong~Loong Fah.
\newblock Active segmentation with fixation.
\newblock In \emph{Computer Vision, 2009 IEEE 12th International Conference
  on}, pages 468--475. IEEE, 2009.

\bibitem[Mishra et~al.(2012)Mishra, Shrivastava, and
  Aloimonos]{mishra2012segmenting}
Ajay~K Mishra, Ashish Shrivastava, and Yiannis Aloimonos.
\newblock Segmenting “simple” objects using rgb-d.
\newblock In \emph{Robotics and Automation (ICRA), 2012 IEEE International
  Conference on}, pages 4406--4413. IEEE, 2012.

\bibitem[Montabone and Soto(2010)]{montabone2010human}
Sebastian Montabone and Alvaro Soto.
\newblock Human detection using a mobile platform and novel features derived
  from a visual saliency mechanism.
\newblock \emph{Image and Vision Computing}, 28\penalty0 (3):\penalty0
  391--402, 2010.

\bibitem[Rao et~al.(2010)Rao, Le, Phoka, Quigley, Sudsang, and
  Ng]{rao2010grasping}
Deepak Rao, Quoc~V Le, Thanathorn Phoka, Morgan Quigley, Attawith Sudsang, and
  Andrew~Y Ng.
\newblock Grasping novel objects with depth segmentation.
\newblock In \emph{Intelligent Robots and Systems (IROS), 2010 IEEE/RSJ
  International Conference on}, pages 2578--2585. IEEE, 2010.

\bibitem[Ren et~al.(2012)Ren, Bo, and Fox]{ren2012rgb}
Xiaofeng Ren, Liefeng Bo, and Dieter Fox.
\newblock Rgb-(d) scene labeling: Features and algorithms.
\newblock In \emph{Computer Vision and Pattern Recognition (CVPR), 2012 IEEE
  Conference on}, pages 2759--2766. IEEE, 2012.

\bibitem[Sch{\"a}fer et~al.(2013)Sch{\"a}fer, Lenzen, and
  Garbe]{schafer2013depth}
Henrik Sch{\"a}fer, Frank Lenzen, and Christoph~S Garbe.
\newblock Depth and intensity based edge detection in time-of-flight images.
\newblock In \emph{3DV}, pages 111--118, 2013.

\bibitem[Tejani et~al.(2014)Tejani, Tang, Kouskouridas, and
  Kim]{tejani2014latent}
Alykhan Tejani, Danhang Tang, Rigas Kouskouridas, and Tae-Kyun Kim.
\newblock Latent-class hough forests for 3d object detection and pose
  estimation.
\newblock In \emph{Computer Vision--ECCV 2014}, pages 462--477. Springer, 2014.

\end{thebibliography}

\end{document}